\documentclass[11pt,a4paper]{article}
\usepackage[hyperref]{acl2021}
\usepackage{times}
\usepackage{latexsym}

\usepackage{microtype}

\aclfinalcopy 

\usepackage{subcaption}
\usepackage{booktabs}
\usepackage{comment}

\usepackage{hyperref}
\usepackage{url}
\usepackage{todonotes}
\usepackage{color,soul}
\usepackage{graphicx}
\usepackage[normalem]{ulem}
\usepackage{mathtools}
\usepackage{algorithmicx}
\usepackage{algpseudocode}
\usepackage{amsmath,amssymb}
\usepackage{hyperref}
\usepackage{url}
\usepackage{todonotes}
\usepackage{graphicx}
\usepackage{microtype}

\usepackage{amsmath,amsfonts,bm}

\def\eqref#1{equation~\ref{#1}}

\def\1{\bm{1}}

\DeclareMathAlphabet{\mathsfit}{\encodingdefault}{\sfdefault}{m}{sl}
\SetMathAlphabet{\mathsfit}{bold}{\encodingdefault}{\sfdefault}{bx}{n}

\def\sD{{\mathbb{D}}}

\def\sT{{\mathbb{T}}}

\newcommand{\reals}{\mathbb{R}}

\newcommand{\zz}{\mathbf{z}}

\newcommand{\modelM}{\mathbb{M}}

\usepackage{url}

\usepackage{xspace}

\title{How transfer learning impacts linguistic knowledge in deep NLP models?}

\author{
Nadir Durrani \hspace{11mm} Hassan Sajjad \hspace{11mm} Fahim Dalvi \\
{\tt \{ndurrani,hsajjad,faimaduddin\}@hbku.edu.qa} \\ 
Qatar Computing Research Institute, HBKU Research Complex, Doha 5825, Qatar \\ 
}

\date{}

\begin{document}
\maketitle
\begin{abstract}

Transfer learning from pre-trained neural language models towards downstream tasks has been a predominant theme in NLP recently. 
Several researchers have shown that deep NLP models learn non-trivial amount of linguistic knowledge, captured at different layers of the model. 
We investigate how fine-tuning towards downstream NLP tasks impacts the learned linguistic knowledge. We carry out a study across popular pre-trained models BERT, RoBERTa and XLNet using layer and neuron-level diagnostic classifiers. We found that for some GLUE tasks, the network relies on the core linguistic information and preserve it deeper in the network, while for others it forgets. Linguistic information is distributed  in the pre-trained language models 
but becomes localized to the lower layers post-fine-tuning, reserving higher layers for the task specific knowledge. The pattern varies across architectures, with BERT retaining linguistic information relatively deeper in the network compared to RoBERTa and XLNet, where it is predominantly delegated to the lower layers.

\end{abstract}

\section{Introduction}

Contextualized word representations learned in transformer-based language models capture rich linguistic knowledge, making them ubiquitous for transfer learning towards downstream NLP problems such as Natural Language Understanding tasks e.g. GLUE 
\cite{wang-etal-2018-glue}. The general idea is to pretrain representations on large scale unlabeled data and adapt these towards a downstream task using supervision.  

Descriptive methods in neural interpretability investigate what knowledge is learned within the representations through relevant extrinsic phenomenon varying from word morphology \cite{vylomova2016word, belinkov:2017:acl,dalvi:2017:ijcnlp} to 
high level concepts such as structure \cite{shi-padhi-knight:2016:EMNLP2016,linzen2016assessing}  and semantics \cite{qian-qiu-huang:2016:P16-11, belinkov:2017:ijcnlp} or more generic properties such as sentence length \cite{adi2016fine,bau2018identifying}. These studies are carried towards analyzing 
representations from pre-trained models. However, it is important to investigate how this learned knowledge evolves 
as the models are adapted towards a specific task 
from the more generic task 
of language modeling \cite{peters-etal-2018-deep} 
that they are primarily trained on. 

In this work, we analyze representations 
of 3 popular pre-trained models (BERT, RoBERTa and XLnet)  with respect to morpho-syntactic and semantic knowledge, as they are fine-tuned towards GLUE tasks. More specifically we investigate i) if the fine-tuned models retain the same amount of linguistic information, ii) how this information is redistributed across different layers and individual neurons. To this end, we use \emph{Diagnostic Classifiers}  \cite{hupkes2018visualisation,conneau2018you}, a popular framework for probing knowledge in neural models. The central idea is to extract feature representations from the network 
and train an auxiliary classifier to predict the property of interest. The quality of the trained classifier on the given task serves as a proxy to the quality of the extracted representations w.r.t to the understudied property \cite{belinkov-etal-2020-linguistic}. 

We carry layer-wise  \cite{liu-etal-2019-linguistic} and neuron-level probing analyses \cite{dalvi:2019:AAAI} to study the fine-tuned representations. The former 
probes representations from individual layers w.r.t a linguistic property and the latter finds 
salient neurons in the network that capture the property. 
Fine-tuning involves adjusting feature weights, therefore it is important to look at the individual neurons to uncover important details, in addition to a more holistic layer-wise view.

Our layer-wise analysis shows: i) that some GLUE tasks rely on core linguistic knowledge and the model preserves the information deeper in the network, while for others it is retained only in the lower layers  
ii) interesting cross-architectural differences 
with knowledge regressed to lower layers in RoBERTa and XLNet as opposed to BERT where it is still retained at the higher layers. Our neuron-wise analysis shows: i) salient linguistic neurons are relocated from the higher to lower layers, reinforcing our layer-wise results, ii) that linguistic information becomes less distributed and less redundant in the network post fine-tuning. 

Finally, we show how our analysis entails findings in layer pruning. 
Dropping higher layers of the models maintains comparable performance to fine-tuning the full network, with linguistic information regressed to the lower layers. 
Conversely, pruning the lower layers (which hold the core linguistic information) leads to substantial degradation in performance.

In comparison to the related work done in this direction, our findings resonate with \newcite{merchant-etal-2020-happens} who found that fine-tuning primarily affects top layers and does not lead to ``catastrophic forgetting of linguistic phenomena'' in BERT. However, we found that other models like RoBERTa and XLNet, which they did not study, see a substantial drop in accuracy even at the lower layers and start forgetting linguistic knowledge much earlier in the network. 
In contrast to \newcite{mosbach-etal-2020-interplay-fine}, 
we study core-linguistic phenomena whereas their study is based on sentence level probing tasks. Differently from both, we carry out a fine-grained neuron analysis which sheds light on how neurons are distributed and relocated post fine-tuning. Our work complements their findings while extending the layer-wise analysis to core-linguistic tasks and additionally looking at the distribution and relocation of neurons 
after fine-tuning.


\begin{figure*}[t]
    \centering
    \begin{subfigure}[b]{0.30\linewidth}
    \centering
    \includegraphics[width=\linewidth]{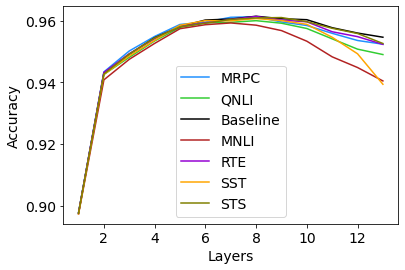}
    \caption{BERT -- POS}
    \label{fig:bert_pos}
    \end{subfigure}
    \begin{subfigure}[b]{0.30\linewidth}
    \centering
    \includegraphics[width=\linewidth]{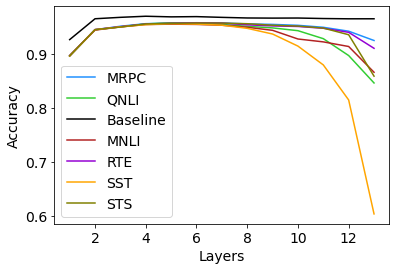}
    \caption{RoBERTa -- POS}
    \label{fig:RoBERTa_pos}
    \end{subfigure}    
    \begin{subfigure}[b]{0.30\linewidth}
    \centering
    \includegraphics[width=\linewidth]{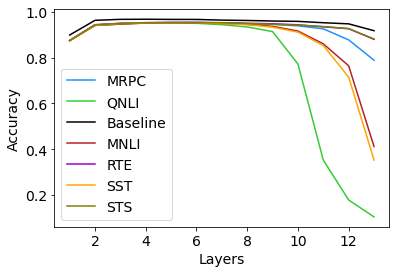}
    \caption{XLNet -- POS}
    \label{fig:xlnet_pos}
    \end{subfigure}
    \begin{subfigure}[b]{0.30\linewidth}
    \centering
    \includegraphics[width=\linewidth]{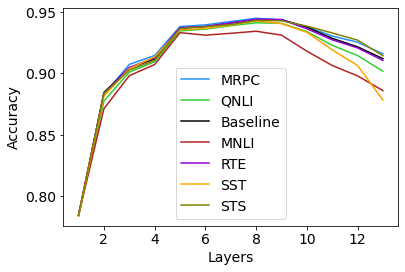}
    \caption{BERT -- Chunking}
    \label{fig:bert_chunking}
    \end{subfigure}
    \begin{subfigure}[b]{0.30\linewidth}
    \centering
    \includegraphics[width=\linewidth]{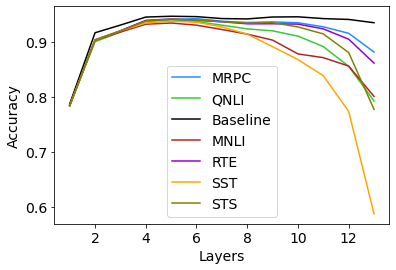}
    \caption{RoBERTa -- Chunking}
    \label{fig:RoBERTa_chunking}
    \end{subfigure}
    \begin{subfigure}[b]{0.30\linewidth}
    \centering
    \includegraphics[width=\linewidth]{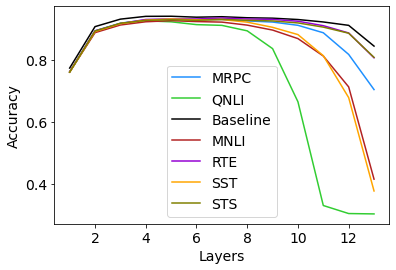}
    \caption{XLNet -- Chunking}
    \label{fig:xlnet_chunking}
    \end{subfigure}

    \caption{Layer-wise Probing Performance. Baseline refers to the performance of the pre-trained models without any finetuning.}
    \label{fig:layerwise}
\vspace{-6pt}
\end{figure*}

\section{Methodology}
\label{sec:methodology}

Our methodology is based on the probing framework called as \emph{Diagnostic Classifiers}. We train a 
classifier using the activations generated from the trained neural network as static features, towards the task of predicting a certain linguistic property. The underlying  
assumption is that if the classifier can predict the property, 
the representations implicitly encode this information. We train layer- and neuron-wise probes using logistic-regression classifiers.  Formally, consider a pre-trained neural language model $\mathbf{M}$ with $L$ layers: $\{l_1, l_2, \ldots, l_L\}$. Given a dataset $\sD=\{w_1, w_2, ..., w_N\}$ 
with a corresponding set of linguistic annotations $\sT=\{t_{w_1}, t_{w_2}, ..., t_{w_N}\}$, we map each word $w_i$ in the data $\sD$ to a sequence of latent representations: $\sD\xmapsto{\modelM}\zz = \{\zz_1, \dots, \zz_n\}$. The model is trained by minimizing the following loss function:
%
%
\begin{equation}
\mathcal{L}(\theta) = -\sum_i \log P_{\theta}(t_{w_i} | w_i) + \lambda_1 \|\theta\|_1 + \lambda_2 \|\theta\|^2_2 \nonumber
\end{equation}
%
where $P_{\theta}(t_{w_i} | w_i) = \frac{\exp (\theta_l \cdot \zz_i)}{\sum_{l'} \exp (\theta_{l'} \cdot \zz_i)} $
 is the probability that word $i$ is assigned property $t_{w_i}$. We extract representations from the individual layers for our layer-wise analysis and the entire network for the neuron-analysis. We use the 
 \emph{Linguistic Correlation Analysis} as described in \newcite{dalvi:2019:AAAI}, to generate a neuron ranking with respect to the understudied linguistic property: Given the trained classifier \mbox{$\theta \in \reals^{D \times T}$}, the algorithm extracts a ranking of the $D$ neurons in the model $\modelM$ based on weight distribution. The elastic-net regularization \cite{Zou05regularizationand} -- a combination of $\lambda_1 \|\theta\|_1$ and $\lambda_2 \|\theta\|^2_2$ is used to strike a balance between identifying focused ($L1$) versus distributed ($L2$) neurons. The weights for the regularization terms are tuned using a grid-search algorithm. 
 
 Following \newcite{durrani-etal-2020-analyzing}, we extract salient neurons for a linguistic property 
 by iteratively 
 choosing the top $N$ neurons from the ranked list and retrain the classifier using these neurons, until the classifier obtains an accuracy close (within a specified threshold $\delta$) to the \emph{Oracle} -- accuracy of the classifier trained using all the features in the network. 

\section{Experimental Setup}
\label{sec:experiments}

\paragraph{Pre-trained Neural Language Models:} We 
experimented with 3 transformer models: BERT \cite{devlin-etal-2019-bert}, RoBERTa \cite{liu2019roberta} and XLNet \cite{yang2019xlnet} using the base versions (13 layers and 768 dimensions). 
This choice of architectures leads to an interesting comparison between auto-encoder versus auto-regressive models. The models were then fine-tuned towards GLUE tasks of which we experimented with SST-2 for sentiment analysis with the Stanford sentiment treebank 
\cite{socher-etal-2013-recursive}, MNLI for natural language inference 
\cite{williams-etal-2018-broad}, QNLI for Question NLI 
\cite{rajpurkar-etal-2016-squad}, 
RTE for recognizing textual entailment 
\cite{Bentivogli09thefifth}, MRPC for Microsoft Research paraphrase corpus 
\cite{dolan-brockett-2005-automatically}, and STS-B for the semantic textual similarity benchmark 
\cite{cer-etal-2017-semeval}. All the models were fine-tuned with the identical settings and we did 3 independent runs. 

\paragraph{Linguistic Properties:} We evaluated our method on 3 linguistic tasks: POS tagging using 
the Penn TreeBank \cite{marcus-etal-1993-building}, syntactic chunking using CoNLL 2000 shared task dataset \cite{tjong-kim-sang-buchholz-2000-introduction}, and semantic tagging using the Parallel Meaning Bank data \cite{abzianidze-EtAl:2017:EACLshort}. We used standard splits for training, development and test data. 

\paragraph{Classifier Settings:} 
We used a linear probing classifier with elastic-net regularization, using a categorical cross-entropy loss, optimized by Adam \cite{kingma2014adam}. Training is run with shuffled mini-batches of size 512 and stopped after 10 epochs. The regularization weights are trained using grid-search. For sub-word based models, we use the last activation value to be the representative of the word 
following \newcite{durrani-etal-2019-one}. We computed \emph{selectivity}  \cite{hewitt-liang-2019-designing} to ensure that our results reflect the property of representations and not the probe's capacity to memorize. Please see Appendix for details.

\section{Analysis}

\subsection{Layer-wise Probing}
\label{sec:layerWise}

First we train layer-wise probes to show how linguistic knowledge is redistributed across the network as we fine-tune it towards downstream tasks. Figure~\ref{fig:layerwise} shows results for POS and Chunking tasks.\footnote{The observations are consistent for semantic tagging. Please see Appendix for results.} We found varying observations across different GLUE tasks. 

\paragraph{Comparing GLUE tasks:} We found that linguistic phenomena are more important for certain downstream tasks, for example STS, RTE and MRPC where they are preserved in the higher layers post fine-tuning, as opposed to others, for example SST, QNLI and MNLI where they are forgotten in the higher layers. It 
would be interesting to study this further by connecting linguistic probes with any causation analysis on these tasks. Such an analysis would shed light on what concepts are used by the network while making predictions and why such information is forgotten for certain tasks. We leave this exploration for future.

\paragraph{Comparing Architectures:} We found that pre-trained models behave differently in preserving information post fine-tuning. In the case of 
BERT, linguistic knowledge is fully preserved 
until layer $9$, 
after which different task-specific models drop to varying degree, with SST and QNLI showing significant drop compared to others. An exception to this overall trend is MNLI 
where we start seeing a decline in performance earlier (between layers $5-7$). Contrastingly RoBERTa and XLNet show a depreciation in linguistic knowledge as early as layer 5. Also the drop is much more catastrophic in these two models with accuracy dropping by more than 35\% in RoBERTa and 70\% in 
XLNet. These results indicate that BERT retains its primarily learned linguistic knowledge and uses only a few of the final layers for fine-tuning, as opposed to XLNet and RoBERTa, where linguistic knowledge is retained only in the lower half of the network. 
Another cross-architectural observation that we made was 
that in RoBERTa and XLNet, the fine-tuned models do not ever reach the baseline performance (i.e. accuracy before fine-tuning -- See Figure \ref{fig:layerwise}) 
at any layer, although the loss is $<2\%$. We conjecture this discrepancy is due to the fact that the knowledge is more redundant and polysemous in the case of BERT, compared to XLNet, where it is more localized (also observed in \newcite{durrani-etal-2020-analyzing}). Consequently, during fine-tuning XLNet and RoBERTa are more likely to lose linguistic information that is unimportant to the downstream task. We discuss this further in our neuron-analysis section.

\begin{figure*}[ht]
    \centering
    \begin{subfigure}[b]{0.30\linewidth}
    \centering
    \includegraphics[width=\linewidth]{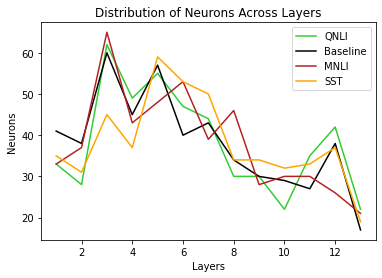}
    \caption{BERT -- SEM}
    \label{fig:bert_pos_selected}
    \end{subfigure}
    \begin{subfigure}[b]{0.30\linewidth}
    \centering
    \includegraphics[width=\linewidth]{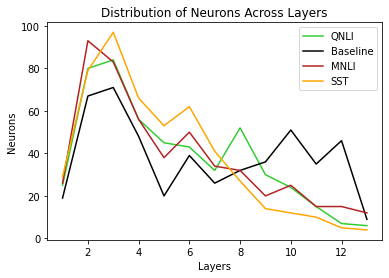}
    \caption{RoBERTa -- SEM}
    \label{fig:roberta_pos_selected}
    \end{subfigure}
    \begin{subfigure}[b]{0.30\linewidth}
    \centering
    \includegraphics[width=\linewidth]{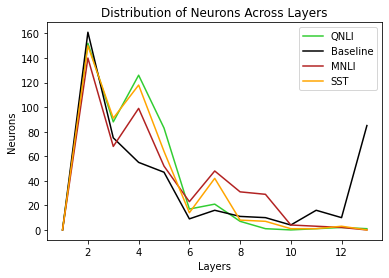}
    \caption{XLNet -- SEM}
    \label{fig:xlnet_pos_selected}
    \end{subfigure}
    \centering
    \begin{subfigure}[b]{0.30\linewidth}
    \centering
    \includegraphics[width=\linewidth]{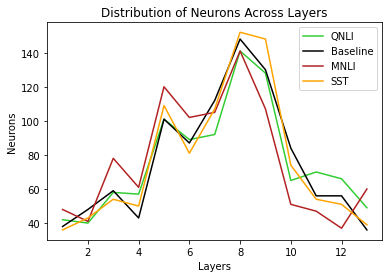}
    \caption{BERT -- Chunking}
    \label{fig:bert_chunking_neurons}
    \end{subfigure}
    \begin{subfigure}[b]{0.30\linewidth}
    \centering
    \includegraphics[width=\linewidth]{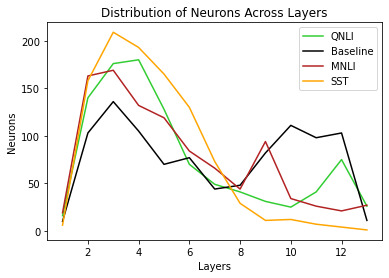}
    \caption{RoBERTa -- Chunking}
    \label{fig:RoBERTa_chunking_neurons}
    \end{subfigure}
    \begin{subfigure}[b]{0.30\linewidth}
    \centering
    \includegraphics[width=\linewidth]{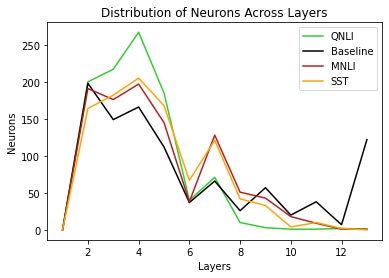}
    \caption{XLNet -- Chunking}
    \label{fig:xlnet_chunking_neurons}
    \end{subfigure}

    \caption{Distribution of top neurons across layers}
    \label{fig:layerwise_neurons_selectedtasks}
\end{figure*}

\subsection{Neuron-wise Probing}
\label{sec:neuronWise}

In our second set of experiments, we conducted analysis at a more fine-grained neuron level using \emph{Linguistic Correlation Method} \cite{dalvi:2019:AAAI}. We extract the most salient neurons w.r.t a linguistic property (e.g. POS) and compare how the distribution of such neurons changes across the network as it is fine-tuned towards a downstream GLUE task. We use the weights of the trained classifier to rank neurons and select minimal set of salient neurons that give the same classifier accuracy as using the entire network in the baseline model. 
We found 5\% neurons for POS and SEM tagging tasks and 10\% for the Chunking tagging were sufficient to achieve the baseline performance.

\paragraph{Information becomes less distributed in the fine-tuned XLNet and RoBERTa models post fine-tuning:} Table~\ref{tab:pos_top_bottom} shows accuracy of the classifier selecting the most (top) and least (bottom) 5\% salient neurons on the task of POS tagging.\footnote{See Appendix for SEM and Chunking tagging.} We observed that the bottom neurons in the fine-tuned models show a significant drop in performance, compared to the baseline model in the case of RoBERTa and XLNet. These results show that the information is more 
redundant in the baseline models as bottom neurons also preserved linguistic knowledge. On the contrary the information becomes more localized and less distributed in the fine-tuned models. The bottom neurons in the fine-tuned BERT changed the least, showing 
that linguistic information is still redundant and distributed in BERT.

\paragraph{How do salient neurons spread across the network layers?} 
Previously we investigated how representations in each layer change w.r.t linguistic task. Now we study how the spread of the most salient neurons changes across the fine-tuned models. Figure \ref{fig:layerwise_neurons_selectedtasks} shows results for the selected GLUE tasks.\footnote{See Appendix for all tasks and linguistic properties.} Notice how the most salient linguistic neurons shift from the higher layers towards the lower layers in 
RoBERTa and XLNet. This is especially pronounced in the case of \emph{Roberta-SST} and \emph{XLNet-QNLI} (See Figures \ref{fig:RoBERTa_chunking} and \ref{fig:xlnet_chunking}), where the number of salient chunking neurons significantly increased in the lower layers and droped in the higher layers, compared to the baseline. 
These findings reinforces our layer-wise results and additionally show how more responsibility is delegated to the neurons in the lower layers. Contrastingly, BERT did not exhibit this behavior. 
These results are inline with \newcite{durrani-etal-2020-analyzing}, who also found linguistic properties in XLNet to be localized to the lower layers\footnote{Similarly \cite{wu-etal-2020-similarity} reported lower and middle layers of XLNet to have the most salient features.} and fewer neurons and mutually exclusive as compared to BERT where neurons are highly polysemous\footnote{attend to multiple linguistic phenomenon} and therefore more redundant.
Their finding helps us explain why XLNet forgets linguistic information that is unimportant to the downstream task more catastrophically.

\begin{table}[]
\centering
\footnotesize
\begin{tabular}{l| rr| rr | rr}
\toprule
 & \multicolumn{2}{c}{\textbf{BERT}} & \multicolumn{2}{c}{\textbf{RoBERTa}} & \multicolumn{2}{c}{\textbf{XLNet}}  \\
\textbf{Tasks} & Top & Bot. & Top & Bot. & Top & Bot. \\ \midrule

Base      & 96.0 & 94.9 & 96.7 & 95.3 & 96.5 & 91.2 \\
\midrule
MRPC      & 95.9 & 94.6 & 95.6 & 91.9 & 95.2 & 78.8   \\
QNLI      & 96.0 & 94.6 & 95.8 & 84.3 & 94.7 & 10.3  \\
MNLI      & 95.8 & 93.9 & 95.4 & 84.8 & 94.9  & 41.1  \\
RTE      & 95.9 & 94.8 & 95.6 & 90.4 & 95.2  & 87.9  \\
SST      & 95.9 & 94.2 & 95.6 & 60.4 & 95.0 & 35.2  \\
STS      & 95.9 & 94.6 & 95.7 & 85.9 & 95.1 & 88.1  \\
\bottomrule
\end{tabular}
\caption{POS accuracy -- Top vs. Bottom neurons}
\label{tab:pos_top_bottom}
\vspace{-2mm}
\end{table}

\section{Network Pruning}
\label{sec:application}

Our layer and neuron-wise analyses showed that core linguistic knowledge is redundant and distributed in the large pre-trained models. But as they are fine-tuned towards a down-stream task, 
it is relocated and localized to lower layers, with higher layers focusing on the task-specific information.
In this section, we show that our findings explain patterns in layer pruning.
We question \textbf{How important is the linguistic knowledge for these downstream NLP tasks?} Following \newcite{sajjad2020poor} we prune top and bottom (excluding the embedding layer) 6 layers of the network in two separate experiments and compare architectures. Table~\ref{tab:layerdrop} shows that removing bottom layers of the network in RoBERTa and XLNet 
leads to more damage compared to BERT. 
\textbf{How do these findings resonate with our analysis?} 
We showed that BERT retains linguistic information even at the higher layers of the model as opposed to RoBERTa where it is preserved predominantly at the lower layers. Removing the bottom 6 layers in RoBERTa leads to a bigger drop because the network is completely deprived of the linguistic knowledge. Linguistic knowledge is more distributed in BERT and preserved at the higher layers also which leads to a smaller drop as it can still access this information. We leave a detailed exploration on this for future. 




\begin{table}[]
\centering
\footnotesize
\begin{tabular}{l| rrr }
\toprule
\textbf{Tasks} & \textbf{SST} & \textbf{MNLI} & \textbf{QNLI} \\ \midrule

& \multicolumn{3}{c}{\textbf{BERT}} \\
Baseline     & 92.4 & 84.0 & 91.1  \\
Prune Top 6      & 90.3 & 81.2 & 87.6   \\
Prune Bottom 6      & 88.1 & 78.4 & 83.7   \\
\midrule
& \multicolumn{3}{c}{\textbf{RoBERTa}} \\
Baseline      & 92.2 & 86.4 & 91.7  \\
Prune Top 6      & 92.0 & 84.4 & 90.0    \\
Prune Bottom 6      & 83.7 & 61.6 & 63.7   \\
\midrule
& \multicolumn{3}{c}{\textbf{XLNet}} \\
Baseline      & 93.9 & 86.0 & 90.4  \\
Prune Top 6      & 92.2 & 83.5 & 88.0    \\
Prune Bottom 6      & 87.5 & 68.1 & 83.0  \\
\bottomrule

\end{tabular}
\caption{Pruning Layers in the Models}
\label{tab:layerdrop}
\vspace{-8mm}
\end{table}


%


\section{Conclusion}

We studied how linguistic knowledge evolves as the pre-trained language models are adapted towards downstream NLP tasks. We fine-tuned three popular models (BERT, RoBERTa and XLNet) towards GLUE benchmark and analyzed representations against core morpho-syntactic knowledge. We used probing classifiers to carry out layer and neuron-wise analyses. Our results showed that morpho-syntactic knowledge is preserved at the higher layers in some GLUE tasks (e.g. STS, MRPC and RTE), while forgotten and only retained at the lower layers in others (MNLI, QNLI and SST). Comparing architectures, we found that BERT retains linguistic knowledge deeper in the network. In the case of RoBERTa and XLNet, the information is only preserved in the middle layers. 
This discrepancy is due to the fact that neurons in BERT are more polysemous and distributed as opposed to XLNet and RoBERTa where they are more localized (towards lower layers) and mutually exclusive. We showed that this difference in architectures, 
entails different patterns as we prune top or bottom layers in the network. Our code is publicly 
as part of the NeuroX toolkit \cite{neurox:aaai19:demo}.



\section*{Ethics and Broader Impact}

For this study, we used existing publicly available data sets while following their terms in the licenses. We do not see any harm or ethical issues resulting from our study and findings.  Our study has implications towards the work on interpreting and analyzing deep models.

\bibliographystyle{acl_natbib}
\bibliography{anthology,acl2021}

\clearpage
\appendix

\section{Appendices}

\begin{figure*}[ht]
    \centering
    \begin{subfigure}[b]{0.30\linewidth}
    \centering
    \includegraphics[width=\linewidth]{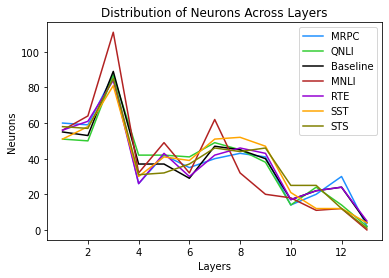}
    \caption{BERT -- POS}
    \label{fig:bert_pos_neurons-A}
    \end{subfigure}
    \begin{subfigure}[b]{0.30\linewidth}
    \centering
    \includegraphics[width=\linewidth]{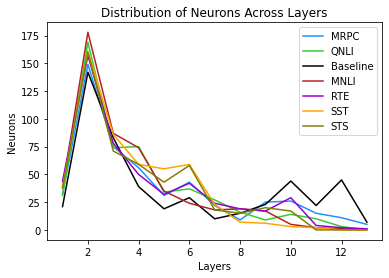}
    \caption{RoBERTa -- POS}
    \label{fig:roberta_pos_neurons-A}
    \end{subfigure}
    \begin{subfigure}[b]{0.30\linewidth}
    \centering
    \includegraphics[width=\linewidth]{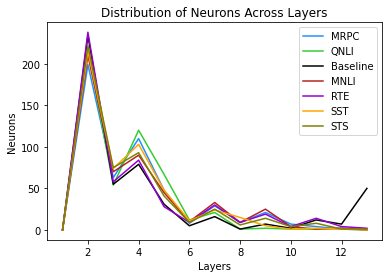}
    \caption{XLNet -- POS}
    \label{fig:xlnet_pos_neurons-A}
    \end{subfigure}
    \centering
    \begin{subfigure}[b]{0.30\linewidth}
    \centering
    \includegraphics[width=\linewidth]{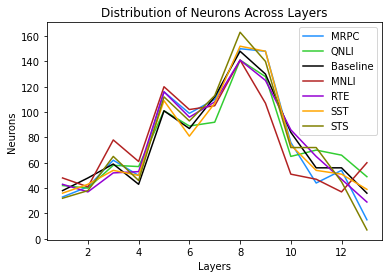}
    \caption{BERT -- Chunking}
    \label{fig:bert_chunking_neurons-A}
    \end{subfigure}
    \begin{subfigure}[b]{0.30\linewidth}
    \centering
    \includegraphics[width=\linewidth]{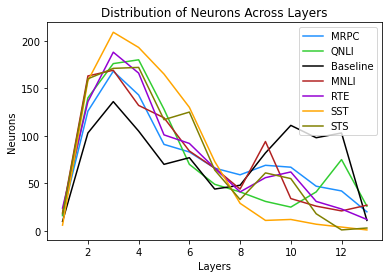}
    \caption{RoBERTa -- Chunking}
    \label{fig:roberta_chunking_neurons-A}
    \end{subfigure}
    \begin{subfigure}[b]{0.30\linewidth}
    \centering
    \includegraphics[width=\linewidth]{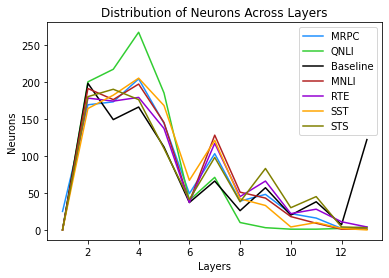}
    \caption{XLNet -- Chunking}
    \label{fig:xlnet_chunking_neurons-A}
    \end{subfigure}
    \centering
    \begin{subfigure}[b]{0.30\linewidth}
    \centering
    \includegraphics[width=\linewidth]{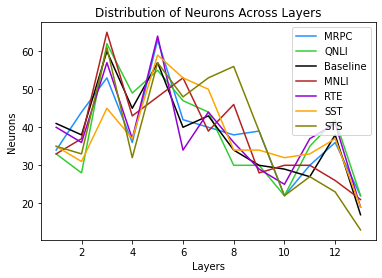}
    \caption{BERT -- SEM}
    \label{fig:bert_sem_neurons-A}
    \end{subfigure}
    \begin{subfigure}[b]{0.30\linewidth}
    \centering
    \includegraphics[width=\linewidth]{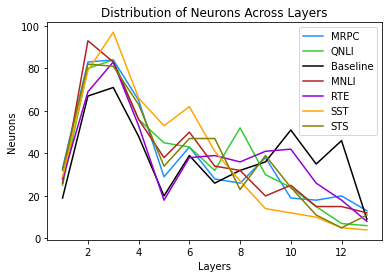}
    \caption{RoBERTa -- SEM}
    \label{fig:roberta_sem_neurons-A}
    \end{subfigure}
    \begin{subfigure}[b]{0.30\linewidth}
    \centering
    \includegraphics[width=\linewidth]{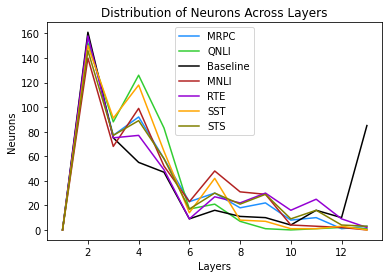}
    \caption{XLNet -- SEM}
    \label{fig:xlnet_sem_neurons-A}
    \end{subfigure}
    \caption{Distribution of Top Neurons across Layers}
    \label{fig:layerwise_neurons_appendix}
\end{figure*}

\begin{figure*}[ht]
    \centering
    \begin{subfigure}[b]{0.30\linewidth}
    \centering
    \includegraphics[width=\linewidth]{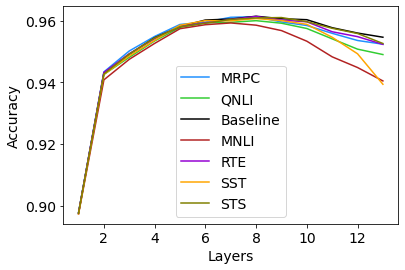}
    \caption{BERT - SEM}
    \label{fig:bert_sem}
    \end{subfigure}
    \begin{subfigure}[b]{0.30\linewidth}
    \centering
    \includegraphics[width=\linewidth]{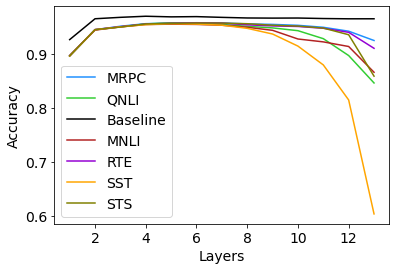}
    \caption{RoBERTa - SEM}
    \label{fig:RoBERTa_sem}
    \end{subfigure}    
    \begin{subfigure}[b]{0.30\linewidth}
    \centering
    \includegraphics[width=\linewidth]{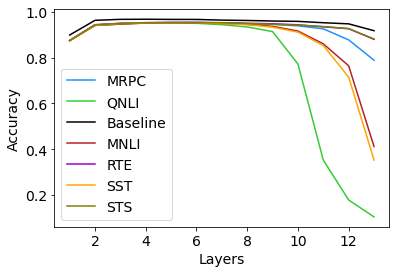}
    \caption{XLNet - SEM}
    \label{fig:xlnet_sem}
    \end{subfigure}
    \caption{Layer-wise Probing Performance}
    \label{fig:layerwise-sem}
\end{figure*}

\subsection{Data and Representations}
\label{subsec:data}

We used standard splits for training, development and test data for the 4 linguistic tasks (POS, SEM, Chunking) that we used to carry out our analysis on. The splits to preprocess the data are available through git repository\footnote{\url{https://github.com/nelson-liu/contextual-repr-analysis}} released with \newcite{liu-etal-2019-linguistic}. See Table \ref{tab:dataStats} for statistics. We obtained the understudied pre-trained models from the authors of the paper, through personal communication.

\begin{table}[ht]									
\centering					
\footnotesize
\resizebox{\columnwidth}{!}{									
    \begin{tabular}{l|cccc}									
    \toprule									
Task    & Train & Dev & Test & Tags \\		
\midrule
    POS & 36557 & 1802 & 1963 & 44 \\
    SEM & 36928 & 5301 & 10600 & 73 \\
    Chunking &  8881 &  1843 &  2011 & 22 \\
    \bottomrule
    \end{tabular}
    }
    \caption{Data statistics (number of sentences) on training, development and test sets using in the experiments and the number of tags to be predicted}

\label{tab:dataStats}						
\end{table}

\subsection{Layer-wise Probing}
\label{subsec:lwProbing}

Section \ref{sec:layerWise} presented layer-wise probing results for POS and Chunking tagging. Figure \ref{fig:layerwise-sem} show results on Semantic tagging. We see a similar pattern across architectures as in Figure \ref{fig:layerwise}.

\subsection{Neuron-wise Probing}
\label{subsec:nwProbing}

Section \ref{sec:neuronWise} presented neuron-wise probing results for for Chunking tagging. Figure \ref{fig:layerwise_neurons_selectedtasks} show results on POS and SEM tagging. We see a similar pattern across architectures as in Figure \ref{fig:layerwise_neurons_appendix}. As the model is fine-tuned towards downstream, number of salient neurons towards a linguistic property, in the lower layers increase.



\subsection{Top versus Bottom Neurons}

In Section \ref{sec:neuronWise} we presented spread how information is more distributed and redundant in in the network as bottom neurons also preserved linguistic knowledge. On the contrary the linguistic information becomes more localized and less distributed post fine-tuning using accuracy of the bottom neurons. Tables~\ref{tab:chunking_top_bottom} and \ref{tab:sem_top_bottom} demonstrate the same pattern with respect to Chunking and Semantic tagging tasks, selecting 10\% and 5\% neurons respectively.

\begin{table}[]
\centering
\footnotesize
\begin{tabular}{l| rr| rr | rr}
\toprule
 & \multicolumn{2}{c}{\textbf{BERT}} & \multicolumn{2}{c}{\textbf{RoBERTa}} & \multicolumn{2}{c}{\textbf{XLNet}}  \\
\textbf{Tasks} & Top & Bot. & Top & Bot. & Top & Bot. \\ \midrule

Base      & 94.7 & 92.3 & 94.8 & 92.5 & 94.2 & 92.3 \\
\midrule
MRPC      & 94.4 & 91.9 & 94.4 & 89.1 & 93.8 & 72.4  \\
QNLI      & 94.3 & 92.3 & 94.0 & 82.2 & 93.0 &  33.3 \\
MNLI      & 93.8 & 91.3 & 93.2 & 82.1 & 92.8  & 44.5   \\
RTE      & 94.7 & 92.2 & 94.3 & 88.2 & 94.0  & 84.7 \\
SST      & 94.3 & 91.9 & 94.1 & 60.7 & 93.8 &  39.7 \\
STS      & 94.8 & 92.3 & 94.3 & 79.7 & 92.2 & 83.8  \\
\bottomrule
\end{tabular}
\caption{Chunking accuracy -- Top vs. Bottom neurons}
\label{tab:chunking_top_bottom}
\end{table}

\begin{table}[]
\centering
\footnotesize
\begin{tabular}{l| rr| rr | rr}
\toprule
 & \multicolumn{2}{c}{\textbf{BERT}} & \multicolumn{2}{c}{\textbf{RoBERTa}} & \multicolumn{2}{c}{\textbf{XLNet}}  \\
\textbf{Tasks} & Top & Bot. & Top & Bot. & Top & Bot. \\ \midrule

Base      & 92.2 & 90.9 & 92.8 & 90.7 & 96.5 & 91.2 \\
\midrule
MRPC      & 92.2 & 90.6 & 91.5 & 88.1 & 92.3 & 72.3   \\
QNLI      & 92.0 & 90.8 & 91.5 & 78.6 & 91.4 & 17.8  \\
MNLI      & 91.9 & 90.3 & 91.4 & 79.0 & 91.3  & 43.0  \\
RTE      & 92.0 & 90.6 & 91.5 & 86.9 & 91.3  & 80.0  \\
SST      & 92.1 & 90.5 & 91.5 & 60.7 & 91.3 & 34.8  \\
STS      & 92.1 & 90.4 & 91.5 & 79.5 & 91.6 & 83.7  \\
\bottomrule
\end{tabular}
\caption{SEM accuracy -- Top vs. Bottom neurons}
\label{tab:sem_top_bottom}
\end{table}

\subsection{Pruning Layers}
\label{sec:pruningLayers}

In Section \ref{sec:application} we showed how pruning bottom layers in RoBERTa was more harmful in comparison to BERT. We conjectured that this pattern entails from our analysis that in RoBERTa linguistic information is preserved in the initial middle layers as opposed to BERT where linguistic knowledge is distributed deeper in the network. We show that XLNet exhibit similar pattern to RoBERTa in Table \ref{tab:layerdrop_appendix}.

\begin{table}[]
\centering
\begin{tabular}{l| rrr }
\toprule
\textbf{Tasks} & \textbf{SST-2} & \textbf{MNLI} & \textbf{QNLI} \\ \midrule

%
& \multicolumn{3}{c}{\textbf{XLNet}} \\
Baseline      & 93.9 & 86.0 & 90.4  \\
Prune Top 6      & 92.2 & 83.5 & 88.0    \\
Prune Bottom 6      & 87.5 & 68.1 & 83.0  \\
\bottomrule

\end{tabular}
\caption{Pruning Layers in the Models}
\label{tab:layerdrop_appendix}
\end{table}

\subsection{Control Tasks} 

While there is a plethora of work demonstrating that contextualized representations encode a continuous analogue of discrete linguistic information, a question has also been raised recently if the representations actually encode linguistic structure or whether the probe memorizes the understudied task.  We use \emph{Selectivity} as a criterion to put a ``linguistic task's accuracy in context with the probe’s capacity to memorize from word types'' \cite{hewitt-liang-2019-designing}. It is defined as the difference between linguistic task accuracy and control task accuracy. An effective probe is recommended to achieve high linguistic task accuracy and low control task accuracy.

\begin{table}[t]									
\centering
\footnotesize
\resizebox{\columnwidth}{!}{
\begin{tabular}{l|ccc}									
\toprule									
 & BERT & XLNet & RoBERTa \\		
 \midrule
 Neu$_a$ & 9984  & 9984 & 9984 \\
\midrule
\multicolumn{4}{c}{POS}  \\
\midrule
Neu$_t$ & 500/5\%  & 500/5\% & 500/5\% \\
Acc$_a$ & 96.2 & 96.4 & 96.3  \\
Acc$_t$ & 95.9 & 96.5 & 96.2 \\
\midrule 
Sel$_a$ & 14.45 & 23.49 & 22.65  \\
Sel$_t$ & 31.68 & 31.82 & 34.21 \\
\midrule
\multicolumn{4}{c}{SEM}  \\
\midrule
Neu$_t$ & 500/5\%  & 500/5\% & 500/5\% \\
Acc$_a$ & 92.51  & 92.29 & 92.95 \\
Acc$_t$ & 92.32 & 92.62 & 92.97  \\
\midrule 
Sel$_a$ & 5.77 & 14.03 & 13.76  \\
Sel$_t$ & 27.17 & 26.55 & 24.53 \\
\midrule
\multicolumn{4}{c}{Chunking}  \\
\midrule
Neu$_t$ & 1000/10\% & 1000/10\% & 1000/10\% \\
Acc$_a$ & 94.36 & 93.84 & 94.66  \\
Acc$_t$ & 94.68 & 94.24 & 94.79  \\
\midrule 
Sel$_a$ & 16.30 & 22.77 & 21.12  \\
Sel$_t$ & 29.19 & 28.42 & 28.91 \\
\bottomrule
\end{tabular}
}
\caption{Selecting minimal number of neurons for each downstream NLP task. Accuracy numbers reported on blind test-set (averaged over three runs) -- Neu$_a$ = Total number of neurons, Neu$_t$ = Top selected neurons, Acc$_a$ = Accuracy using all neurons, Acc$_t$ = Accuracy using selected neurons after retraining the classifier using selected neurons, Sel = Difference between linguistic task and control task accuracy when classifier is trained on all neurons (Sel$_a$) and top neurons (Sel$_t$).
}							
\label{tab:accuracy}		
\end{table}

\subsection{Infrastructure and Run Time}
\label{subsec:runTime}

Our experiments were run on NVidia GeForce GTX TITAN X GPU card. Grid search for finding optimal lambdas is expensive when optimal number of neurons for the task are unknown. Running grid search would take $\mathcal{O} (M N^2)$ where $M = 100 $ (if we try increasing number of neurons in each step by 1\%) and $N = 0, 0.1, \dots 1e^{-7}$. We fix the $M=20\%$ to find the best regularization parameters first reducing the grid search time to $\mathcal{O} (N^2)$ and find the optimal number of neurons in a subsequent step with $\mathcal{O} (M)$. The overall running time of our algorithm therefore is $\mathcal{O} (M + N^2)$. This varies a lot in terms of wall-clock computation, based on number of examples in the training data, number of tags to be predicted in the downstream task. Including a full forward pass over the pre-trained model to extract the contextualized vector, and running the grid search algorithm to find the best hyperparameters and minimal set of neurons took on average 8 hours ranging from 3 hours for the Chunking experiment to 12 hours for POS and SEM due to large training data.

\subsection{Hyperparameters}
\label{subsec:hyperparameters}

We use elastic-net based regularization to control the trade-off between 
selecting focused individual neurons versus group of neurons while maintaining the original accuracy of the classifier without any regularization. We do a grid search on $L_1$ and $L_2$ ranging from values $0 \dots 1e^{-7}$. See Table \ref{tab:lambdas} for the optimal values for each task across different architectures.

\begin{table}[ht]									
\centering					
\resizebox{\columnwidth}{!}{									
    \begin{tabular}{l|ccc}									
    \toprule									
    & BERT & XLNet & RoBERTa  \\		
    \midrule
    \multicolumn{4}{c}{L1 , L2   = $\lambda_1$, $\lambda_2$ }  \\
    \toprule
    POS & .001, .01 & .001, .01 & .001, .001  \\
    SEM & .001, .01 & .001, .01 & .001, .001  \\
    Chunk & $1e^{-4}, 1e^{-5}$  & $1e^{-4}, 1e^{-4}$ & .001, .001 \\
    \bottomrule
    \end{tabular}
    }
    \caption{Best elastic-net lambdas parameters for each task}
\label{tab:lambdas}						
\end{table}

\end{document}